\documentclass[default]{sn-jnl}
\usepackage{moreverb}
\usepackage{hyperref}
\usepackage{amsfonts}
\usepackage{amsmath}
\usepackage{algpseudocode}
\usepackage{algorithm}
\usepackage{subcaption}
\newcommand\BibTeX{{\rmfamily B\kern-.05em \textsc{i\kern-.025em b}\kern-.08em
T\kern-.1667em\lower.7ex\hbox{E}\kern-.125emX}}

\usepackage{graphicx}%
\usepackage{multirow}%
\usepackage{amsmath,amssymb,amsfonts}%
\usepackage{amsthm}%
\usepackage{mathrsfs}%
\usepackage[title]{appendix}%
\usepackage{xcolor}%
\usepackage{textcomp}%
\usepackage{manyfoot}%
\usepackage{booktabs}%
\usepackage{algorithm}%
\usepackage{algorithmicx}%
\usepackage{algpseudocode}%
\usepackage{listings}%

\theoremstyle{thmstyleone}%
\theoremstyle{thmstyletwo}%

\theoremstyle{thmstylethree}%

\begin{document}

\title[Article Title]{Transformer-Driven Modeling of Variable Frequency Features for Classifying Student Engagement in Online Learning}


\author*[1, 3]{\fnm{Sandeep} \sur{Mandia}}

\author[1]{\fnm{Kuldeep} \sur{Singh}}

\author[1]{\fnm{Rajendra} \sur{Mitharwal}}

\author[2]{\fnm{Faisel} \sur{Mushtaq}}

\author[4]{\fnm{Dimpal} \sur{Janu}}

\affil*[1]{\orgdiv{Department of Electronics \& Communication Engineering}, \orgname{Malaviya National Institute of Technology Jaipur},  \postcode{302017}, \state{Rajasthan}, \country{India}}

\affil[2]{\orgdiv{Department of Computer Science \& Engineering}, \orgname{Malaviya National Institute of Technology Jaipur},  \postcode{302017}, \state{Rajasthan}, \country{India}}

\affil[3]{\orgdiv{Department of Electronics \& Communication Engineering}, \orgname{Thapar Institute of Engineering and Technology, Patiala},  \postcode{147004}, \state{Punjab}, \country{India}}

\affil[4]{\orgdiv{Department of Electronics \& Communication Engineering}, \orgname{SRM University AP, Amarawati},  \postcode{522502}, \state{Andhra Pradesh}, \country{India}}



\abstract{The COVID-19 pandemic and the internet's availability have recently boosted online learning. However, monitoring engagement in online learning is a difficult task for teachers. In this context, timely automatic student
engagement classification can help teachers in making adaptive adjustments to meet students’ needs. This paper proposes EngageFormer, a transformer based architecture with sequence pooling using video modality for engagement classification. The proposed architecture computes three views from the input video and processes them in parallel using transformer encoders; the global encoder then processes the representation from each encoder, and finally, multi layer perceptron (MLP) predicts the engagement level. A learning centered affective state dataset is curated from existing open source databases. The proposed method achieved an accuracy of $63.9$\%, $56.73$\%, $99.16$\%, $65.67$\%, and $74.89$\%  on Dataset for Affective
States in E-Environments (DAiSEE), Bahçeşehir University Multimodal Affective Database - 1 (BAUM-1), Yawning Detection Dataset (YawDD), University of Texas at Arlington Real-Life Drowsiness Dataset (UTA-RLDD), and curated learning-centered affective state dataset respectively. The achieved results on the BAUM-1, DAiSEE, and YawDD datasets demonstrate state-of-the-art performance, indicating the superiority of the proposed model in accurately classifying affective states on these datasets. Additionally, the results obtained on the UTA-RLDD dataset, which involves two-class classification, serve as a baseline for future research. These results provide a foundation for further investigations and serve as a point of reference for future works to compare and improve upon.}




\maketitle

\section{Introduction}\label{section:Intro}
The global COVID-19 pandemic has led to the worst crisis including the closure of educational institutions worldwide, affecting millions of students. A UNESCO report revealed that schools in nearly 190 nations and educational institutions in 61 countries had to close down as a result of the pandemic \cite{mahmood2021instructional}. This has resulted in the absence of physical education for 156 million students worldwide, prompting the adoption of virtual platforms like Zoom, WebEx, and Teams for online education \cite{dias2020deeplms}. The popularity of online learning has been bolstered by the widespread availability of the internet, which has made it possible for students to access learning materials from far-off locations and has popularized Massive Open Online Courses (MOOCs). Online education is a cost-effective alternative to the traditional mode of  education in terms of transportation and lodging costs \cite{dhawan2020online}. However, the lack of socialization and interaction with teachers in virtual classrooms has led to increased student withdrawal rates. Thus, providing pedagogical assistance to individual students based on their involvement has become crucial in online learning \cite{adnan2020online}. Automatically measuring student engagement can be a promising way to detect student involvement and facilitate personalized pedagogical assistance, not only in online education but also in other learning environments such as educational games, authentic classrooms, and intelligent tutoring systems\cite{dewan2019engagement}.

\par The engagement is considered to be multi-faceted with behavioral, emotional, and cognitive facets. Behavioral engagement deals with active participation and involvement in learning activities. On the other hand, emotional engagement deals with students' affective state, and cognitive engagement is related to students' zeal and self-dedication to learning and mastering the concept \cite{fredricks2004school}. Students' emotional engagement in learning activities is indirectly related to cognitive engagement \cite{jung2018learning}. Students with  elevated emotional and cognitive engagement are likely to be persistent and put more effort into learning to meet the course outcomes. Student engagement relates to facial expression, upper-body posture, and other environmental factors. However, facial expression is the most studied and promising indicator of engagement level \cite{dewan2019engagement}. Ekman's six basic emotions, happiness, sorrow, anger, surprise, disgust, and contempt, are the most studied emotional states in the affective computing domain \cite{saurav2021emnet}. However, basic emotions are rare in learning environments, whereas curiosity, frustration, boredom, confusion, happiness, and anxiety are frequent affective states \cite{calvo2010affect, mandia2023recognition}. Also, the timing and intensity of these affective states are closely related to engagement in synchronous learning environments \cite{grafsgaard2013automatically}. 

\par The automatic student engagement detection can be categorized into machine learning-based and deep learning-based methods. Machine learning-based methods typically rely on the extraction of hand-crafted features from the data to train a classifier (support vector machines (SVM), or decision trees) to recognize different states of engagement \cite{whitehill2014faces}. On the other hand, deep learning-based methods utilize a series of nonlinear processing layers in between input and output layers to learn complex patterns in the data. These methods can automatically extract high-level features from raw input data, eliminating the need for hand-crafted feature engineering. Deep learning models such as convolutional neural networks (CNNs) and recurrent neural networks (RNNs) have shown promising results in various applications of student engagement detection. \cite{liao2021deep}. The computer vision-based deep learning methods are unobtrusive and easy to implement \cite{dewan2019engagement}. The deep learning-based methods can further be divided into static image-based methods, and video-based methods \cite{saurav2021emnet}. The video-sequence-based data is modeled using different deep learning algorithms, three-dimensional convolution neural network (3D-CNN), CNN with Long Short Term Memory (LSTM) hybrid network, and a combination of these with attention has been explored in the past \cite{gupta2016daisee, liao2021deep, mehta2022three, huang2019fine, mandia2023automatic,abedi2021improving}.

\par Vision-based student engagement recognition has predominantly been accomplished through the use of CNN-based deep learning architectures. CNNs have shown to be highly effective in numerous computer vision tasks. In contrast, transformers are the standard architecture for natural language processing (NLP), utilizing attention mechanisms to incorporate long-range data dependencies \cite{devlin2018bert}. Notably, transformer architectures have demonstrated state-of-the-art performances on various vision tasks \cite{yan2022multiview}, which has motivated researchers to explore their application in vision-based student engagement recognition \cite{mandia2022vision}. In this paper, we propose a robust transformer-based architecture for detecting student engagement. In addition, we present a curated learning-centered affective state dataset, which has been compiled from publicly available datasets, as such datasets are relatively scarce.

\par The main contributions of this work are as follows:
\begin{itemize}
\addtolength{\itemindent}{0cm}
\item This study introduces EngageFormer, a novel three-view transformer model for student engagement classification. It utilizes video modality and combines view and global encoders to capture both slowly and rapidly varying temporal features. EngageFormer incorporates sequence pooling for informative token-level summarization, resulting in accurate classification of student engagement levels.
\item This study also curates a learning-centered affective state dataset from existing public databases, which includes Bored, Confusion, Engaged, Frustration, Sleepy, and yawning affective states.
\item This study also evaluates the effectiveness of EngageFormer model by experimenting on five affective state datasets. The experimental results reflect that the proposed method outperforms the existing state-of-the-art methods on BAUM-1s, YawDD, DAiSEE datasets and produces a new baseline for UTA-RLDD dataset on two class classification.
\end{itemize}
The remaining paper is organized as follows. Section \ref{sec:Related work} details related work for automatic student engagement detection. Subsequently, section \ref{sec:methodology} presents the proposed methodology using three view transformer with sequence pooling. The section \ref{Sec:Experiments} outlines the details of datasets used and presents a comparative analysis with the performance of the existing techniques. Finally, the paper is concluded in section \ref{sec:conclusion}.

\section{Related work}\label{sec:Related work}
Traditionally, Self-reports, observational checklists and rating scales, and automatic recognition are the three categories into which the student involvement recognition techniques are divided \cite{whitehill2014faces}. The former two categories do not suit the online environment as they heavily rely on students or external observers and have a coarser temporal resolution \cite{d2006predicting, o2010development}. On the contrary, automatic student engagement recognition using sensory data does not require direct and continuous input from students. Therefore, automated engagement recognition becomes the most relevant 
 subject and natural choice for researchers. Student engagement analysis is studied based on their actions in online learning  (from log data) \cite{5518758,aluja2019measuring}. Although, these techniques depend on limited evaluation metrics and develop biases when used in the wild. 
 \par Automated student engagement recognition is also explored using various physiological sensor signals such as EEG, blood pressure, heart rate, and galvanic skin response \cite{monkaresi2016automated, fairclough2006prediction, khedher2019tracking}. However, the physiological sensors are obtrusive and costly, which makes it difficult to use for student engagement recognition. On the other hand, computer vision-based methods due to their unobtrusive nature are the most popular for student engagement recognition. In particular, face-based methods have been explored largely as the face depicts an emotional state that is intuitively related to engagement. Among the preliminary studies Grafsgaard et al. \cite{6681424} utilized Computer Expression Recognition Toolbox (CERT) for the estimation of more frequent Action Units (AUs) based on the AUs activation. Bosch et al. \cite{bosch2015automatic} estimated students' affective states based on AUs, orientation, and position features of the face, whereas, Saneiro et al. \cite{saneiro2014towards} exploited 2D facial points, 3D head poses, and movement along with AUs for student engagement. Although the AU-based methods can estimate the emotional state of students, they are less robust in the wild environment, thereby making it tedious to predict the detailed engagement of students. In another aspect, appearance-based features have been extracted and fed to the machine learning classifiers for feature learning and classification. Whitehill et al. \cite{whitehill2014faces} first annotated samples for four engagement levels and then trained several classifiers using different facial features. Kamath et al. \cite{kamath2016crowdsourced} used SVM based on facial Histogram of Oriented Gradient (HOG) features to enhance the prediction performance. Monkaresi et al. \cite{monkaresi2016automated} used an ensemble of features derived from AUs, Local Binary Pattern on Three Orthogonal Planes (LBP-TOP), and heart rate to train the automatic student engagement detector. Although the algorithm generalizes well with hand-crafted features. however, designing hand-crafted features for different environmental conditions is cumbersome.

Deep learning's overwhelming performance in computer vision tasks has motivated the research community to use it for automatic student engagement recognition. Among the initial studies, Nezami et al. \cite{mohamad2020automatic} utilized a multilayer CNN trained on a custom dataset for student engagement recognition. In another study, Schulc et al. \cite{schulc2019automatic} proposed a CNN-LSTM-based model to predict if a person is attentive or non-attentive. The aforementioned studies utilized a small private dataset to model a deep neural network and thereby making it difficult to make a comparison between their performances and intrinsic challenges. To overcome such a challenge and set a benchmark for comparative evaluation among the state-of-the-arts for student engagement detection, Gupta et al. \cite{gupta2016daisee}  proposed a database namely Dataset for Affective States in E-Environment (DAiSEE) with five baselines for engagement estimation. Among the first, Zhang et al. \cite{zhang2019novel} utilized DAiSEE to perform binary classification and achieved the highest accuracy on student engagement classification based on Inflated 3D convolution (I3D). Similarly, Huang et al. \cite{huang2019fine} proposed a Deep Engagement Recognition Network (DERN) trained on DAiSEE for engagement classification. Their study achieved an accuracy of  60\% on four class engagement predictions. 
Another dataset was proposed by Dhall et al. \cite{dhall2019emotiw} called EmotiW  is widely used in the field of affective computing for student engagement prediction. Several studies have been carried out utilizing EmotiW dataset to estimate student engagement including the study of Niu et al. \cite{niu2018automatic}. Niu et al. proposed a GRU-based model trained on novel Gaze-AU-Pose (GAP) features. A similar study was conducted by Yang et al. \cite{yang2018deep} which used an LSTM model as a multi-model regressor  and outperformed all the previous benchmarks on the EmotiW dataset. To improve the detection performance,  Lio et al. \citep{liao2021deep} and Mehta et al. \cite{mehta2022three} incorporated attention modules in addition to employing deep learning models in order to focus on the most relevant features responsible for student engagement. Whereas the former employed Se-ResNet-50 and LSTM for spatial and temporal feature learning, the latter utilized 3D DenseNet to model engagement using DAiSEE and EmotiW datasets. Although most of these methods have shown satisfactory performance on an individual dataset, their robustness can not be proved. Furthermore, the student engagement video datasets are very few in number and small in scale.

Most of the existing methods use CNNs for spatial feature extraction which suffer due to local receptive field and image-specific inductive bias \cite{dosovitskiy2020image}.
 On the contrary, transformers learn global dependencies by employing a module called self-attention. The self-attention induces the importance of other frames into the one being processed. Furthermore, it is hard to estimate beforehand at what frequency and duration the affective states will appear in the learning context \cite{dewan2019engagement}. Considering the limitations of CNNs and the innate challenges intrinsic to student engagement estimation. We propose EngageFormer, a pure multi-view transformer \cite{yan2022multiview} based architecture for spatial and temporal modeling to efficiently estimate student engagement from the videos. The EngageFormer overcomes the problem of modeling the affect states appearing at different frequencies and duration by capturing the slowly and rapidly varying features using three views of the proposed model.
 
\section{Methodology}\label{sec:methodology}
In this section, we will model the student engagement classification problem from video modality. The data points of student learning online are represented as video input and corresponding label pairs $(\textbf{X}, y)$, Where $\textbf{X}$ is a sequence of RGB frames of students attending online classes represented as:
\begin{equation}
    \mathbf{X} = \{\mathbf{X}_{t} ; t = 1\cdot\cdot\cdot T\} 
    \label{Eq:Equation}
\end{equation}
The scalar $y$ is the output label for the video. The task is a six-label classification problem, i.e., $y\in\{1....C\}$, for $C=6$. 
\begin{figure*}
    \centering
    \includegraphics[width=12cm,height=4cm]{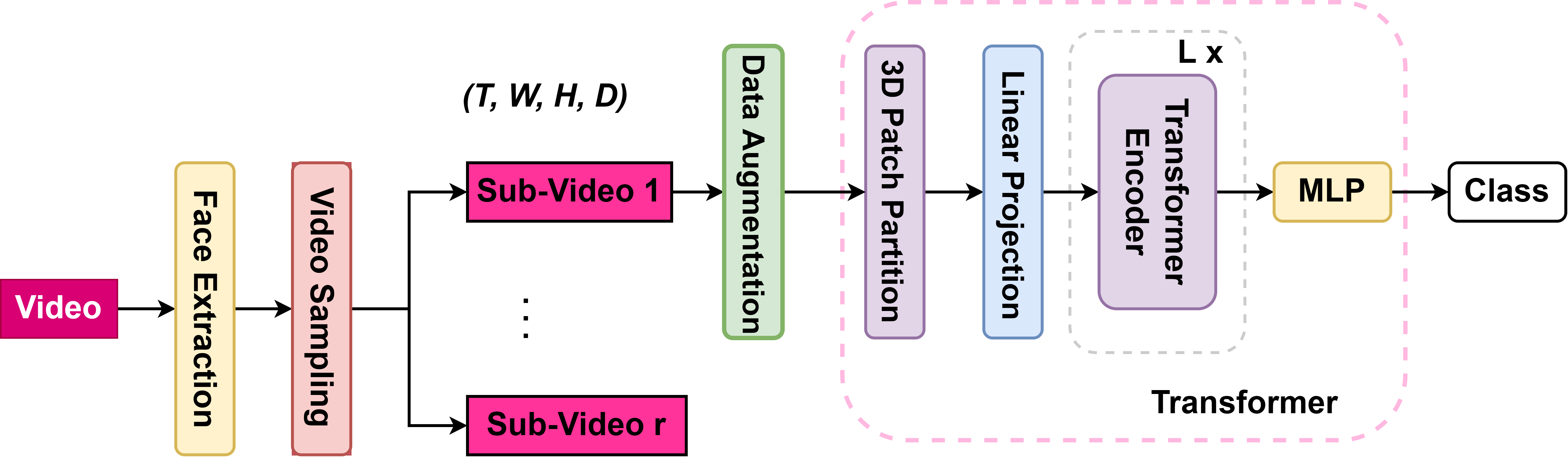}
    \caption{Proposed student engagement classification methodology. Where $L$ is the number of encoder layers and $T, W, H,W$ are dimensions of the pre-processed video.}
    \label{fig1:methodology}
\end{figure*}

The overall methodology for automatic student engagement estimation is shown in Fig. \ref{fig1:methodology}. As the face is the most expressive region for describing the affective states responsible for student engagement. Therefore, the input video is first pre-processed by the Multi-task Cascaded Convolutional Network (MTCNN) \cite{zhang2016joint}  to extract faces from raw input video frames. The video is then sampled at different rates, and $r$ sub-videos are then extracted from the pre-processed face video and provided as input to the proposed transformer model. Random augmentation techniques, as discussed in Sec. \ref{Subsec:Experimental setup}, are eventually applied to the input video for better model generalization. Finally, a three-view transformer, which is capable of modeling slowly and rapidly varying features in the temporal axis is employed to extract robust spatial features and perform student engagement classification. In addition, an attention-based sequence pooling method which accounts for information from all tokens is utilized to represent the transformer output. Finally, the output class is predicted using the MLP unit of the architecture shown in fig. \ref{fig1:methodology}.
\begin{figure}
    \centering
    \includegraphics[width=7cm,height=8cm]{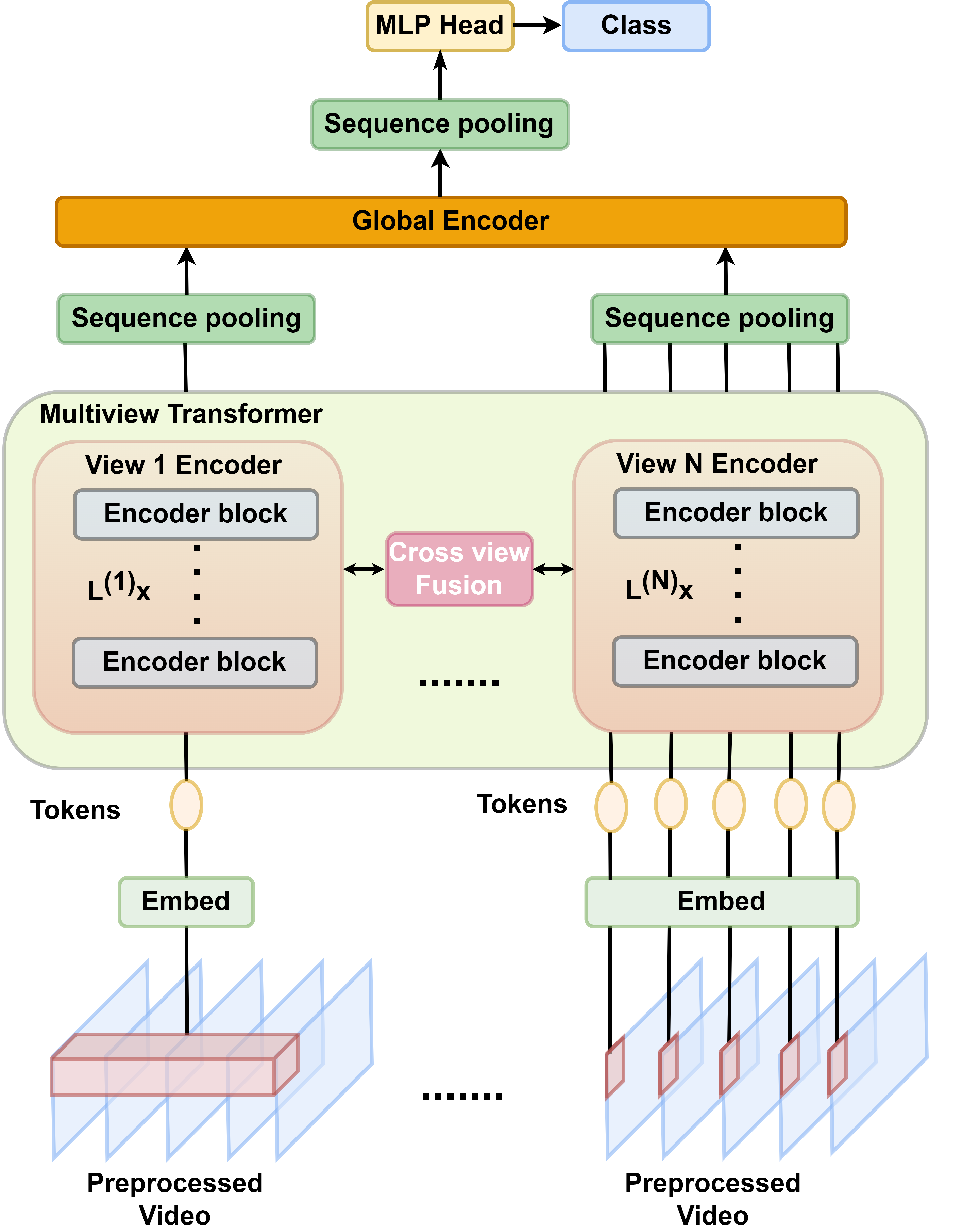}
    \caption{Proposed transformer architecture}
    \label{fig1:architecture}
\end{figure}
To explain the detailed proposed architecture shown in Fig. \ref{fig1:architecture}, we begin with preliminaries vision transformer (ViT), and its video counterpart Video Vision Transformer (ViViT). Subsequently, the tokenization process is explained and finally each detail of proposed transformer architecture is presented.

\subsection{Preliminaries}
A video sample is presented as $X\in\mathbb{R}^{T\times H\times W\times D}$, where $T$, $H$, $W$, and $D$ correspond to the temporal, height, width, and depth of the input volume. This input volume is processed by the transformer by first computing discrete tokens and successively feeding them into various encoder layers of the transformer architecture to produce a compact representation of input by learning global features.

The first transformer-based model for vision task, ViT \cite{dosovitskiy2020image} extracts tokens by linearly projecting non-overlapping patches from image. For videos, these tokens are computed as described in ViViT by extracting Spatio-temporal "tubes", $x_1, x_2, x_3, ..., x_N \in \mathbb{R}^{t\times h\times w\times c}$ from the input video $X$, where $N=\lfloor\frac{T}{t}\rfloor\times \lfloor\frac{H}{h}\rfloor\times \lfloor\frac{W}{w}\rfloor$ \cite{arnab2021vivit}. A linear operator $E$ is applied to project each tube $x_i$ to a token $z_i \in \mathbb{R}^d$, as $z_i = Ex_i$. All the tokens are concatenated, and a learnable class token $z_{cls}\in \mathbb{R}^d$ \cite{devlin2018bert} is prepended. Since transformers are order agnostic, a sequence is added with the position embedding $P \in \mathbb{R}^{(N+1)\times d}$. Therefore, the tokenization process can be represented as:
\begin{equation}
    z^0 = [z_{cls},Ex_1, Ex_2, Ex_3, ..., Ex_N] + P
    \label{Eq:Equation1}
\end{equation}

Typically, a 3D convolution is used to accomplish the linear projection, with the kernel size being $t \times h \times w$ and the strides being $(t, h, w)$ in the time, height, and width dimensions, respectively. Then, $L$ transformer encoder layers process the token sequence in $z$. Each layer, $l$ is applied sequentially and can be represented as:
\begin{equation}
    y^l = MSA(LN(z^{l-1})) + z^{l-1}
    \label{Eq:Equation2}
\end{equation}
\begin{equation}
    z^l = MLP(LN(y^{l})) + y^{l}
    \label{Eq:Equation3}
\end{equation}
Where MSA is multi-head self-attention \cite{vaswani2017attention}, LN is layer normalization \cite{ba2016layer}, and MLP is multi-layer perceptron with two layers separated by GeLu \cite{hendrycks2016gaussian} activation function. Finally, an MLP classifier, $W^{out}\in \mathbb{R}^{d\times C}$ maps the cls token $z_{cls}$ to one of the $C$ classes.
\subsection{Tokenization}
The input to the transformer is given in the form of tokens. As shown in Fig. \ref{fig1:architecture}, first we construct multiple views $z^{0,(1)}, z^{0,(2)}, ..., z^{0,(V)}$ from input video by obtaining Spatio-temporal tubelets of variable dimensions as tokens. Where the number of views are represented by $V$ and $z^{l,(i)}$ represents output of $l^{th}$ layer of $i^{th}$ view. A view is a representation of video in a set of fixed-size tubelets. A larger view comprises larger tubelets and fewer tokens, and vice-versa. We have utilized 3D convolution for tokenization in the proposed architecture, as shown in Fig. \ref{fig1:architecture}. Different convolution kernels were used for each view. As reported in \cite{yan2022multiview}, smaller tubelets capture the fine-grained details, whereas larger tubelets incorporate slowly varying semantics of the expression. We used separate transformer encoders for processing each view extracted from the input video.
\subsection{Transformer}\label{Sub:transformer}
Since the self-attention mechanism of the transformer has quadratic computational complexity \cite{vaswani2017attention}, which makes processing all tokens concurrently infeasible. Therefore, the extracted tokens $Z^0 = [z^{0,(1)}, z^{0,(2)}, ...., z^{0,(V)}]$ from multiple views are processed using multiple transformer encoders each having  $L^{i}$ transformer layers as shown in Fig. \ref{fig1:architecture}. To propagate the information from one view to another, lateral (cross) connections are made. Finally, the representation of each view is computed using sequence pooling by applying an attention-based operation. Representations of all views are further processed using a global encoder with sequence pooling at the output. 
\subsubsection{Encoder}\label{SubSub:Encoder}
\begin{figure}
    \centering
    \includegraphics[width=7cm,height=8cm]{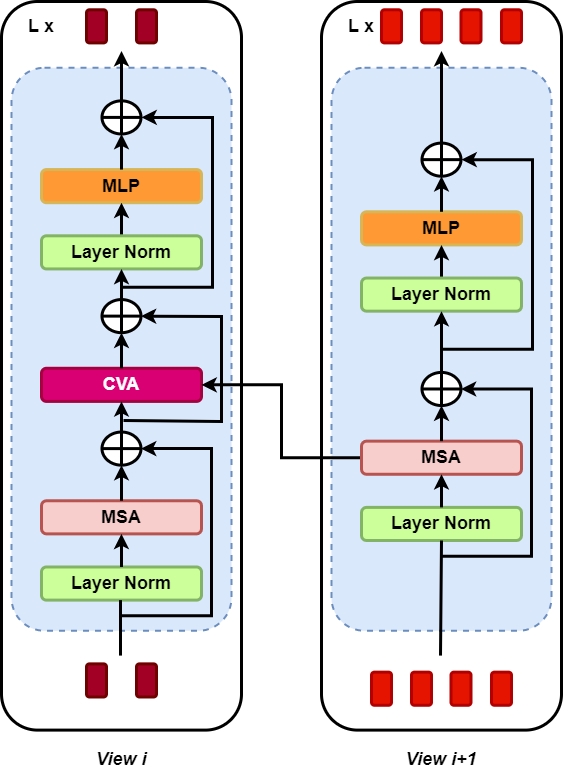}
    \caption{Encoder with cross-view attention fusion}
    \label{fig2:encoder}
\end{figure}

In the proposed study each view is processed by an individual transformer encoder and the information between the views is fused by using lateral (cross) connections as shown in Fig. \ref{fig2:encoder}.
Each encoder layer within the transformer architecture follows the original design proposed by Vaswani et al. \cite{vaswani2017attention} except that the lateral connections for multi-view information fusion are made as described in Sec. \ref{SubSub:CVAF} and sequence pooling at the output of the encoder as described in Sec. \ref{subsec:Sequence pooling}. Furthermore, unlike multiview transformer\cite{yan2022multiview}, we compute self-attention among all the tokens of a view to compute the output representation. 
\subsubsection{Cross view attention fusion}\label{SubSub:CVAF}
As mentioned, we fuse information among views. We use cross-view attention fusion (CVAF) to fuse information among views, as it was the best fusion method reported by Yan et al. \cite{yan2022multiview}. The views are first ordered in ascending order of the number of tokens (\emph{i.e.} $N^{(i)} \leq N^{(i+1)}$), and the information is fused among all pairs of two adjacent views, $i$ and $i+1$. The tokens of larger view, $z^{(i)}$ are updated by computing attention between view $i$ and view $i+1$ by taking $z^{(i)}$ as queries and $z^{(i+1)}$ as keys and values. The keys and values are projected to the same dimension as that of the queries, and CVAF is computed as follows
\begin{equation}
    z^{(i)} = CVAF(z^{(i)}, W^{proj}z^{(i+1)}) + z^{(i)},
    \label{Eq:Equation4}
\end{equation}
\begin{equation}
    CVAF(x,y) = Softmax \left( \frac{W^QxW^Ky^T}{\sqrt{d_k}}\right)W^Vy. 
    \label{Eq:Equation5}
\end{equation}
Here $W^Q$, $W^K$, and $W^V$ are query- key- and value-weight matrices used in attention during training process. As shown in Fig. \ref{fig2:encoder} and denoted in Eq. \ref{Eq:Equation4}, a residual connection around the cross-view attention fusion operation is also applied. 
\begin{table*}[t]
\centering
\caption{Model configuration for view and global encoders}
\label{Tab1:Model configuration}
\begin{tabular}{cccccc}
\\ \hline
Module Name     & \begin{tabular}[c]{@{}c@{}}Hidden \\ Size \end{tabular}& \begin{tabular}[c]{@{}c@{}} MLP \\ dimension \end{tabular} & \begin{tabular}[c]{@{}c@{}}Number of \\ attention heads\end{tabular} & \begin{tabular}[c]{@{}c@{}}Number of \\ encoder layers\end{tabular}  \\ \hline
View Encoder   & 512& 1024 & 3  & 3 \\
Global Encoder & 512 & 1024 & 5  & 1 \\ \hline     
\end{tabular}
\end{table*}
\subsection{Sequence pooling}\label{subsec:Sequence pooling}
In order to aggregate the information from each view, Yan et al. \cite{yan2022multiview} follow the ViT \cite{dosovitskiy2020image} to extract the class token from all views. On the contrary, we use sequence pooling, first proposed by Hassani et al. \cite{hassani2021escaping}, to extract representation from each view of the proposed model architecture. The utilization of sequence pooling stems from the understanding that information is dispersed throughout all video segments and, therefore, across all tokens within the sequence. Consequently, it becomes necessary to aggregate this information by assigning appropriate weight to each token. Sequence pooling involves employing an attention-based approach to transform the sequence of tokens into a vector representation. This enables the summation and consolidation of the information encapsulated within the tokens, facilitating subsequent analysis and processing of the sequence. The sequence pooling can be seen as a transform $T : \mathbb{R}^{N^{(i)} \times d} \mapsto \mathbb{R}^{d}$. Given the output $z^{(i)} \in \mathbb{R}^{N^{(i)}\times d}$ from $i^{th}$ view, the token weights are calculated by applying a linear weight matrix $W^S \in \mathbb{R}^{d \times 1}$ and Softmax activation to the token sequence as shown in Eq. \ref{Eq:Equation6}:
\begin{equation}
    Weights = Softmax([z^{(i)}W^S]^T) \in \mathbb{R}^{1\times N^{(i)}}
    \label{Eq:Equation6}
\end{equation}
The Eq. \ref{Eq:Equation7} assign weights to each input token based on contextual information, semantic relationships, and relevant features. These weights reflect the tokens' relative importance in capturing the sequence's underlying dynamics. The calculated weights are applied to the tokens adjusting their contributions employing a weighting operation as follows:
\begin{equation}
    R_p^{(i)} = Weights Z^{(i)} \in \mathbb{R}^{1 \times d} 
    \label{Eq:Equation7}
\end{equation}
The $R_p^{(i)}$ is the vector representation of view $i$. Vector representations of all views are aggregated to be provided as an input to the global encoder.
\subsection{Global encoder}\label{Subsec:Global encoder}
As depicted in Fig. \ref{fig1:architecture}, the representation tokens derived from each view are fed into the global encoder, facilitating the integration of information from all views. The global encoder efficiently combines the representation tokens from multiple views to capture a comprehensive understanding of the underlying information.
At the output of the global encoder, a sequence pooling operation, as elaborated in Sec. \ref{subsec:Sequence pooling}, is subsequently employed to generate the video representation. This sequence pooling operation aggregates the token representations, capturing the essential information from the sequence. Finally, the resulting representation token obtained from the global encoder is mapped to one of the $C$ output labels, enabling classification based on the learned features.

We have used the "view transformer encoder" for processing each view and a global transformer encoder to process the representations from all views. The cross view attention fusion is used to fuse the information among different views, and the sequence pooling is used to compute the encoder representation. After global encoder, an MLP finally classifies the student engagement.
\setlength{\arrayrulewidth}{0.3mm} 
\setlength{\tabcolsep}{0 pt} 
\renewcommand{\arraystretch}{1}
\begin{table*}[h!]
\centering
\caption{Comparison of results on BAUM-1s dataset}
\label{tab:results on BAUM-1s}
\begin{tabular}{|l|lccc|} \hline
\textbf{Ref} & \textbf{Method} & \textbf{Acc.(\%)}&\textbf{P}&\textbf{R} \\ \hline \hline
    Zhalehpour et al., 2016 \cite{zhalehpour2016baum}& SVM with LPQ features&45.04&-&- \\
    Zhang et al., 2017 \cite{zhang2017learning}& C3D with SVM & 50.11&-&-\\
    Ma et al., 2019 \cite{ma2019audio}&3D CNN, DBN, SVM&54.69&-&- \\
    Pan et al., 2021 \cite{pan2021multimodal}&ELM&55.38&-&-\\
    Proposed method&EngageFormer& \textbf{56.73}&51.99&47.07\\
    \hline
\end{tabular}
\end{table*}
\begin{figure*}[t]
    \begin{subfigure}{.5\textwidth}
        \includegraphics[width=.8\linewidth,height=5cm]{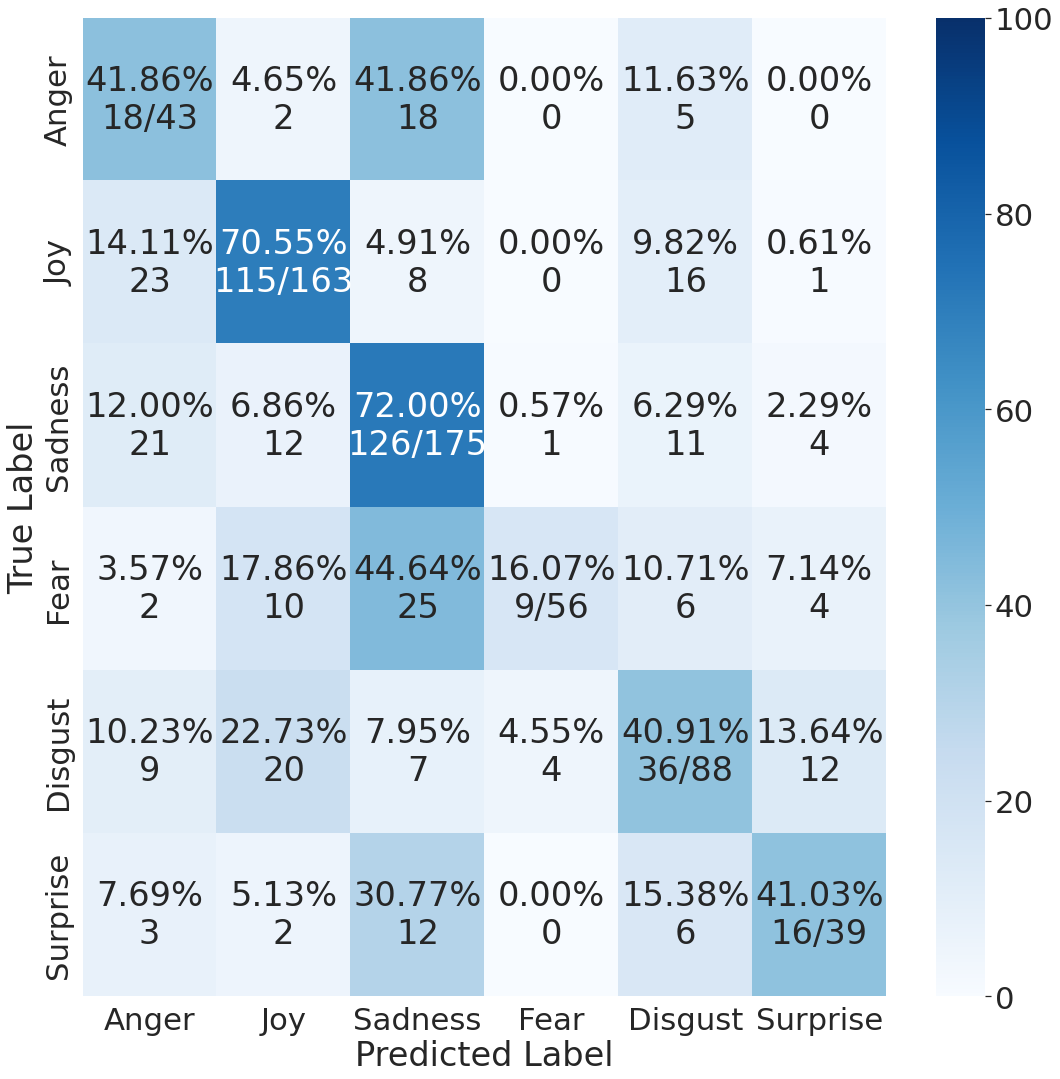}
        \subcaption{}
        \label{fig:Confusion matrix BAUM-1s}
    \end{subfigure}
    \begin{subfigure}{.5\textwidth}
        \includegraphics[width=.8\linewidth,height=5cm]{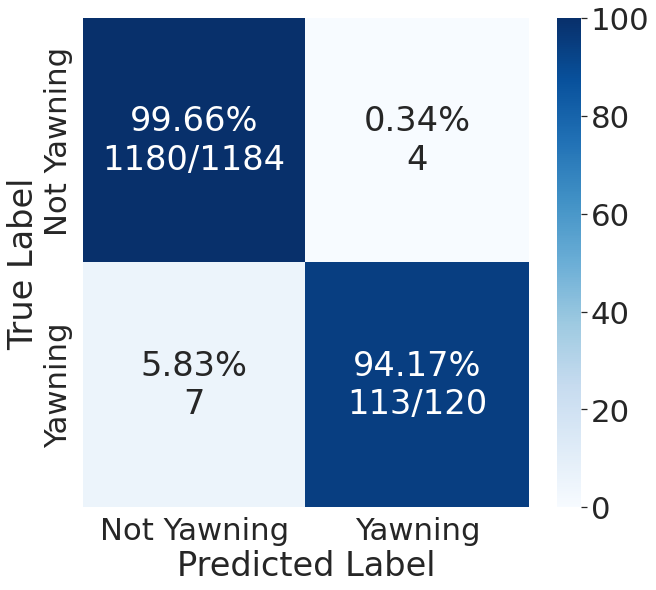}
        \subcaption{}
        \label{fig:Confusion matrix YawDD}
    \end{subfigure}
    \caption{Confusion matrix for classification on (a) BAUM-1s dataset and (b) YawDD dataset}
\end{figure*}
\section{Experiments}\label{Sec:Experiments}
\subsection{Experimental setup}\label{Subsec:Experimental setup}
In this subsection, we present the experimental setup in terms of model hyperparamters, and architecture modules. We used the same transformer encoder for processing each view and a different global transformer encoder architecture to aggregate the information from all views. The settings of the view and global encoders are presented in Table \ref{Tab1:Model configuration}.
\setlength{\arrayrulewidth}{0.3mm} 
\setlength{\tabcolsep}{4.5pt} 
\renewcommand{\arraystretch}{1}
We used three-view model with tubelets of sizes $2 \times 8\times 8$, $4 \times 8\times 8$, and $8 \times 8\times 8$, first being the temporal dimension. The model is trained for $100$ epochs on video samples of 32 frames temporal length. To optimize the network an AdamW \cite{loshchilov2017decoupled} optimizer with an initial learning rate of  $1e^{-4}$ (with cosine decay) and weight decay of $1e^{-5}$ is used. The input image spatial resolution is set to $112 \times 112$ for both training and inference since the face is the only region of interest in this study. In addition, we have also used some data augmentation and regularization schemes such as label smoothing, stochastic depth, Gaussian noise, and vertical flip to train the model efficiently \cite{huang2016deep,szegedy2016rethinking}.
\subsection{Datasets}\label{SubSec:Datasets}
The efficacy of the proposed model is assessed by conducting evaluations on a diverse collection of video datasets detailed below specifically designed for affective state classification. These datasets encompass a wide range of emotional states and cover various contexts, ensuring the model's ability to generalize across different affective states and scenarios. Through rigorous evaluation on these datasets, the performance and effectiveness of the proposed model in accurately classifying affective states can be thoroughly examined and quantitatively measured.
\subsubsection{DAiSEE}\label{SubSubsec:DAiSEE}
The DAiSEE \cite{gupta2016daisee} dataset contains 10-second long 9068 videos recorded from 112 persons at $1920\times 1080$ resolution. The videos are recorded at 30 fps with a full HD webcam in unconstrained environments while students watched educational tutorials on a computer screen. The video snippets are annotated by experts into engaged, bored, confused, and frustrated affective states. The affective states are further four intensity levels namely, (1) very low, (2) low, (3) high, (4) very high. In our experiments, we exclusively make use of the engaged affective state.
\setlength{\arrayrulewidth}{0.3mm} 
\setlength{\tabcolsep}{1 pt} 
\renewcommand{\arraystretch}{1}
\begin{table*}[h!]
\centering
\caption{Comparison of results on YawDD dataset} \label{Tab:Results on YawDD dataset}
\begin{tabular}{|l|lccc|} \hline
\textbf{Ref} & \textbf{Method} & \textbf{Acc (\%)} & \textbf{P} & \textbf{R} \\ \hline \hline
Omidyeganeh et al.,2016 \cite{omidyeganeh2016yawning}& Modified Viola-Jones&75&-&- \\ 
    Zang et al., 2017 \cite{zhang2017driver}         &   LSTM              &    88.6&-&87.1                 \\
     Zang et al., 2015  \cite{zhang2015driver}      &   CNN              &      92&-&- \\
     Bai et al., 2021 \cite{bai2021two} & CNN+GCN &93.4&85.5&94.0 \\
     Deng and Wu, 2019 \cite{deng2019real} & Multiple CNN with KCF&96.3&- &-\\ 
     Ji et al., 2019 \cite{ji2019fatigue}&Mouth State Recognition-Net&98.45&-&-\\
     Ye et al., 2021 \cite{ye2021driver} & RCAN &98.43&98.78&98.53\\
     Xiang et al., 2022 \cite{xiang2022driving}&3D-SE-Net(T)&99.03&-&-\\
     Proposed&EngageFormer&$\mathbf{99.16}$&96.58&94.17 \\ \hline
\end{tabular}
\end{table*}
\subsubsection{BAUM-1}\label{SubSubsec:BAUM1}
The BAUM-1 \cite{zhalehpour2016baum} is an audio-visual dataset including mental and affective states. It consists of BAUM-1a(acted) and BAUM-1s (spontaneous) databases. The BAUM-1s database is used for experiments, it includes eight affective states (joy, anger, sadness, disgust, fear, surprise, boredom, contempt) and four mental states (unsure, thinking, concentrating, bothered). The BAUM-1s database includes 1222 video snippets recorded at $720\times 576$ resolution from 31 Turkish persons. We use six basic emotions for our experiments. 
\subsubsection{UTA-RLDD}\label{SubSubSec:RLDD}
Ghoddoosian et al. \cite{ghoddoosian2019realistic} proposed the university of Texas at Arlington Real-Life Drowsiness Dataset (UTA-RLDD) database for multi-stage drowsiness detection. The database includes micro-expressions under fatigue cases along with extreme and easily observable features. The videos are recorded by sixty participants for $30$ h using their cell phones and webcams in the real life. The dataset contains 180 videos, each approximately 10 minutes long. The videos are labeled for alert, low vigilant, and drowsy classes.
\subsubsection{YawDD}\label{SubSubSec:YAWDD}
The Yawning Detection Dataset (YawDD) \cite{abtahi2014yawdd} contains two sub-datasets recorded from two locations of the camera. The first is recorded from a camera fitted beneath the car's front mirror, it includes 322 videos with and without glasses/sunglasses of both male and female subjects. The second is recorded from the camera mounted on the dashboard of the car, 29 videos are recorded with similar conditions. The videos are recorded at $640 \times 480$ resolution, 30 fps, and stored in AVI format. The annotation is done for various activities such as normal, talking, yawning, and a mix of these. The video clips are closely cropped to elect the yawning activity and converted into a two-class dataset for our experiments.
\subsubsection{Learning centered affective state dataset}\label{subSubSec:curated dataset}
\setlength{\arrayrulewidth}{0.1mm} 
\setlength{\tabcolsep}{7 pt} 
\renewcommand{\arraystretch}{1}
\begin{table}[t]
\centering
\caption {Summary of curated dataset where B:Boredom, C:Confusion, F:Frustration, E:Engaged, S:Sleepy, Y:Yawning} 
\label{tab:curated dataset}
\begin{tabular}{|l|llllll|}
                \hline
                & \textbf{B} & \textbf{C} & \textbf{F} & \textbf{E} & \textbf{S} & \textbf{Y} \\ \hline
\textbf{DAiSEE} & 327              & 79                 & 73                   & 327              & -               & -                \\
\textbf{YawDD}  & -                & -                  & -                    & -                & -               & 217              \\
\textbf{BAUM-1} & 22               & -                  & -                    & 62               & -               & -                \\
\textbf{UTA-RLDD}   & -                & -                  & -                    & -                & 62              & -                \\\hline
\textbf{Total}  & 349              & 79                 & 73                   & 389              & 62              & 217 \\ \hline 

\end{tabular}
\end{table}
Ekman's basic emotions \cite{ekman1992argument} are the most studied in Facial Expression  Recognition (FER) domain. However, these basic emotions are not frequent and related to learning goals \cite{pekrun2000social}. Therefore, we curate a learning-centered affective state dataset from existing open-source databases. Bian et al. \cite{bian2019spontaneous} proposed confusion, distraction, fatigue, enjoyment, and neutral relevant affective states in online learning, and Ashwin \& Guddeti \cite{ashwin2019unobtrusive} considered boredom, confused, engaged, frustrated, sleepy, and neutral as learning-centered affective states and reported a student engagement study.  The learning-centered affective state dataset includes Boredom, Confusion, Engaged, Frustration, Sleepy, and Yawning affective states included from DAiSEE, YawDD, BAUM-1, and UTA-RLDD datasets. The learning-centered dataset contains $1169$ videos samples (Boredom:349, Confusion:79, Frustration:73, Engaged:389, Sleepy:62, and Yawning:217) the details are presented in Table \ref{tab:curated dataset}. For training and testing purposes, the dataset is split into an $80:20$ ratio. The proposed methodology is also evaluated on a learning-centered affective state dataset. 
\setlength{\arrayrulewidth}{0.3mm} 
\setlength{\tabcolsep}{5 pt} 
\renewcommand{\arraystretch}{1}
\begin{table*}[h!]
\centering
\caption{Fold-wise results on RLDD dataset with confusion matrix (CM)}
\label{tab:results on RLDD dataset}
\begin{tabular}{|l|llllll|} \hline
\textbf{Fold}     & \textbf{Acc(\%)} &\textbf{P}&\textbf{R}& \textbf{}             & \multicolumn{2}{c|}{\textbf{CM}} \\ 
  &                                        & \multicolumn{1}{l}{}  &&& \multicolumn{2}{l|}{Predicted}   \\ \hline \hline
\multirow{2}{*}{} Folder 1 in test set& \multirow{2}{*}{}     68.11&66.33&71.78& \multirow{2}{*}{\rotatebox[origin=c]{90}{True}} &262                & 103               \\
                  & &&& &132 &240 \\ \hline
\multirow{2}{*}{} Folder 2 in test set& \multirow{2}{*}{}     65.26&71.89&48.12& \multirow{2}{*}{\rotatebox[origin=c]{90}{True}} &179                & 193               \\
                  & &&& &70 &302 \\ \hline               
\multirow{2}{*}{} Folder 3 in test set& \multirow{2}{*}{}     64.72&59.10&91.67& \multirow{2}{*}{\rotatebox[origin=c]{90}{True}} &341                & 31               \\
                  &&& & &236 &136 \\ \hline
\multirow{2}{*}{} Folder 4 in test set& \multirow{2}{*}{}     64.45&60.75&79.73& \multirow{2}{*}{\rotatebox[origin=c]{90}{True}} &291                & 74               \\
                  &&& & &188 &184 \\ \hline
\multirow{2}{*}{} Folder 5 in test set& \multirow{2}{*}{}     65.8&67.17&60.55& \multirow{2}{*}{\rotatebox[origin=c]{90}{True}} &221                & 144               \\
                  &&& & &108 &264 \\ \hline
\end{tabular}
\end{table*}

\setlength{\arrayrulewidth}{0.3mm} 
\setlength{\tabcolsep}{2 pt} 
\renewcommand{\arraystretch}{1}
\begin{table*}[h!]
\centering
\caption{Comparison of results on DAiSEE dataset}
\label{tab:results on DAiSEE}
\begin{tabular}{|l|lccc|}
\hline
\textbf{Ref} & \textbf{Method} & \textbf{ Acc(\%)}&\textbf{P}&\textbf{R} \\ \hline \hline

    Gupta et al., 2016\cite{gupta2016daisee}&Frame level InceptionNet&47.1&-&-\\
    Gupta et al., 2016\cite{gupta2016daisee} &Video level InceptionNet&46.4&-&- \\
    Jianfei et al.,2018\cite{yang2018deep}&I3D&52.4&-&- \\
    Gupta et al., 2016\cite{gupta2016daisee} & C3D Fine Tuning&56.1&-&- \\
    Gupta et al., 2016\cite{gupta2016daisee}&LRCN&57.9&-&- \\
    Liao et al., 2021\cite{liao2021deep}&DFSTN&58.8&35.5&19.26 \\
    Huang et al., 2019 \cite{huang2019fine}&DERN&60.0&-&- \\
    Mehta et al., 2022\cite{mehta2022three}&3D DenseAttNet&63.59&35.31&36.50 \\
    Proposed method&EngageFormer&\textbf{63.9}&35.64&46.38\\     \hline
\end{tabular}
\end{table*}
\begin{figure*}[h!]
    \begin{subfigure}{.5\textwidth}
        \includegraphics[width=.8\linewidth, height=5cm]{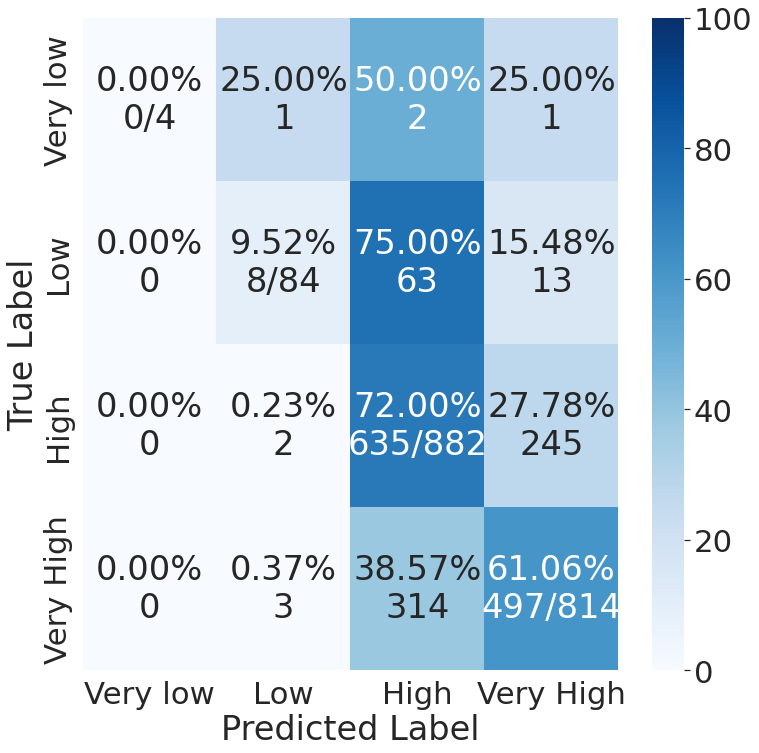}
        \subcaption{}
        \label{fig:Confusion matrix DAiSEE}
    \end{subfigure}
    \begin{subfigure}{.5\textwidth}
        \includegraphics[width=.8\linewidth, height=5cm]{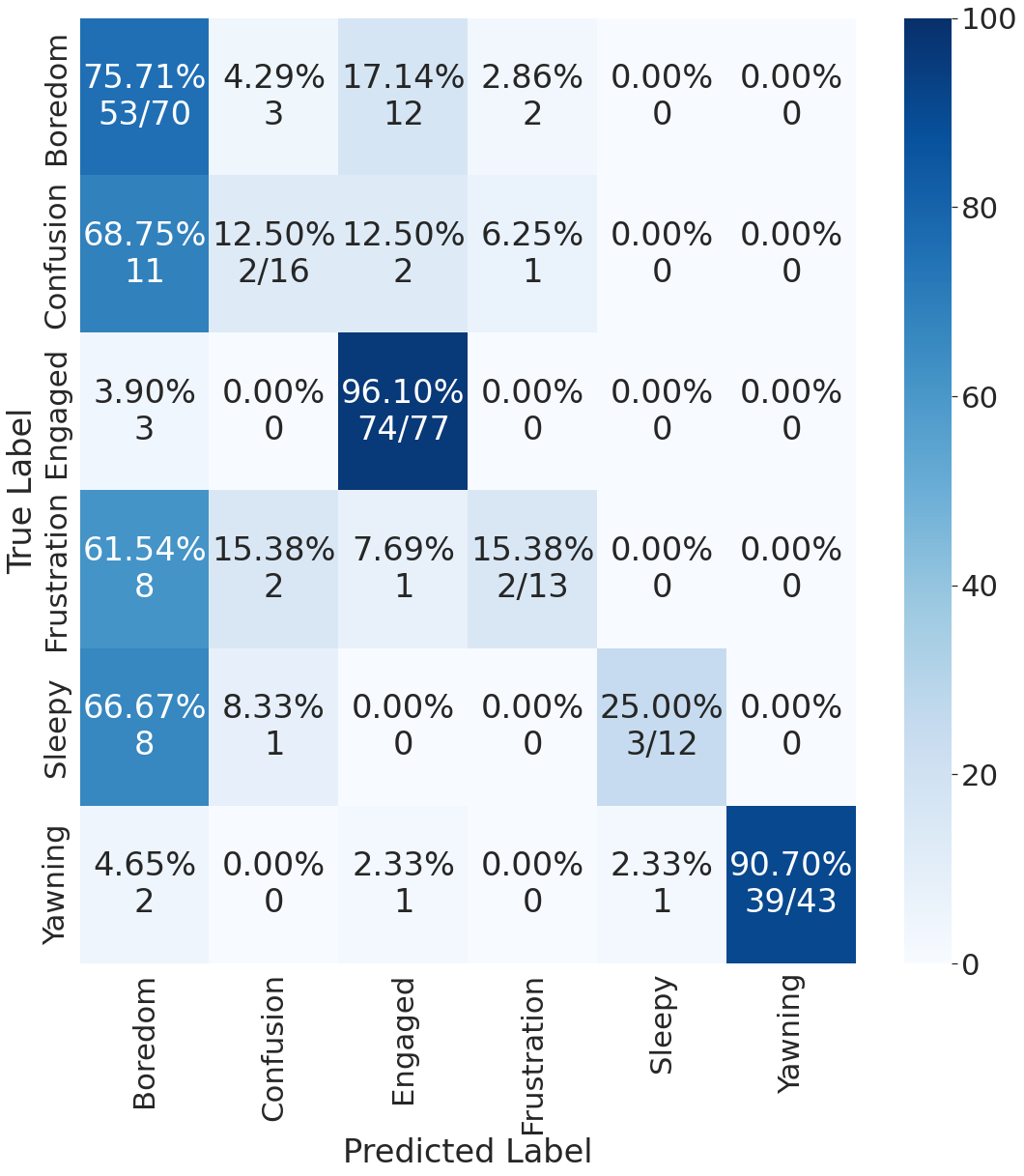}
        \subcaption{}
        \label{fig:Affective state dataset}
    \end{subfigure}
    \caption{Confusion matrix for (a) DAiSEE dataset (b) Learning centered affective state dataset}
\end{figure*}
\subsection{Experimental results}\label{SubSec:Results}
In this subsection, results on above mentioned open-source datasets are discussed.
The results on six basic emotions (anger, joy, sadness, fear, disgust, and surprise) of the BAUM-1s dataset using only vision modality are shown in Table \ref{tab:results on BAUM-1s}. The boldface shows the best result, as one can notice, the proposed method achieved the best performance with a $56.73\%$ accuracy score. The proposed method achieved 51.99\% precision (P) and 47.07\% recall (R). Fig. \ref{fig:Confusion matrix BAUM-1s} shows the confusion matrix of the classification of six basic emotions.
The results on the YawDD dataset are shown and compared with the existing works in Table \ref{Tab:Results on YawDD dataset}. It can be observed From Table \ref{Tab:Results on YawDD dataset} that the proposed method shows the highest accuracy. Furthermore, Fig. \ref{fig:Confusion matrix YawDD} presents the confusion matrix of classification of the YawDD dataset. Considering Yawning a positive class, the confusion matrix shows that precision is greater than recall therefore the proposed method prefers detection over false alarm. 
To evaluate the proposed method on the RLDD dataset, we considered only the alert and drowsy states. Five-fold-cross validation is performed by keeping one folder in the test and remaining in the training set as proposed in the paper \cite{ghoddoosian2019realistic}. The results are presented in Table \ref{tab:results on RLDD dataset}, accuracy for each fold is in the second column and the confusion matrices with alert being first and drowsy being second class  are shown in column three. The accuracy score ranges from $64.45$ to $68.11$ \%, achieving an average accuracy of $65.67\%$ The proposed model is a classification model which classifies the affective states and we were only interested in determining how much the drowsy class is separable from alert class, we report the two class classification result and our method can be compared with other similar works in future.
The results on the test set of the DAiSEE dataset for the proposed methodology  are compared to the existing works in Table \ref{tab:results on DAiSEE}. The bold font represents the best score for the experiment for four-level engagement prediction. As can be observed, the proposed method has shown improvement over the previous studies thus setting a new benchmark for the future studies in this domain. The confusion matrix for engagement prediction on the DAiSEE dataset is shown in Fig. \ref{fig:Confusion matrix DAiSEE}, similar to previous methods the proposed method could not classify the minority classes well. Finally, we report results on the test set of the learning-centered affective state dataset. The confusion matrix is shown in Fig. \ref{fig:Affective state dataset}, the proposed method achieved $74.89$\% accuracy. It can be noted that the proposed methodology is able to recognize the boredom, engaged, and yawning states with comparatively better accuracy whereas the remaining three with poor accuracy. We believe it is due to fewer data in these affective states during training.

\section{Conclusion}\label{sec:conclusion}
This paper presents a transformer-based automatic student engagement classification method in online learning using video modality. The video is first pre-processed to generate different views of varying video tublets and then given to the three view transformer architecture with sequence pooling to generate the corresponding output representation for each view. Subsequently, the representations from all views are fed to global transformer encoder to aggregate information and finaly the affect state is predicted using an MLP. A learning centered affective state dataset is curated from existing open source databases. The proposed methodology is evaluated on engaged affective state from the DAiSEE dataset and the curated learning-centered affective state dataset and achieved $63.9$\% accuracy on DAiSEE which is higher than the existing results and $74.89$\% accuracy on learning-centered affective state dataset. The proposed methodology is also evaluated on BAUM-1s, YawDD, and UTA-RLDD affective state datasets. The results show that our method outperforms existing works on BAUM-1s and YawDD datasets and therefore sets a new baseline for drowsiness detection on the UTA-RLDD dataset. Future research can use our results for comparison on the UTA-RLDD dataset. To increase the quality of engagement prediction, which is now constrained by the availability of small training sets, a larger dataset will be required in the future. It is also possible to assess the proposed method's applicability in various learning contexts, such as the classroom.

\backmatter








\section*{Declarations}

\subsection*{Funding}
No funding was received for conducting this study.

\subsection*{Conflicts of interests}
The authors declare that they have no conflict of interest.

\subsection*{Data Availability}
Data sharing is not applicable to this article, as open-source datasets were used for analysis in this study.





\bibliography{snarticle.bib}
\bibliographystyle{ieeetr}

\end{document}